\documentclass[lettersize,journal]{IEEEtran}
\usepackage{amssymb,amsmath}
\usepackage{cite}
\pdfoutput=1
\usepackage{graphicx}
\usepackage[caption=false]{subfig}
\usepackage[linesnumbered,ruled,lined]{algorithm2e}
\usepackage{psfrag}
\usepackage{url}
\usepackage[latin1]{inputenc}
\usepackage[absolute,overlay]{textpos}
\usepackage{tikz}
\usetikzlibrary{arrows,calc,decorations.markings}
\usetikzlibrary{topaths}
\usetikzlibrary{shapes,trees}
\usetikzlibrary{arrows}
\usetikzlibrary{shadows}
\usetikzlibrary{positioning}
\usetikzlibrary{matrix}
\usetikzlibrary{shapes.geometric}
\usetikzlibrary{decorations.pathmorphing}
\usepgflibrary{patterns}
\usetikzlibrary{calc}
\usetikzlibrary{fit}					

\usepackage{pgf}
\usetikzlibrary{arrows,automata}
\usepackage[latin1]{inputenc}

\usepackage{textcomp}
\usepackage{xcolor}
\def\BibTeX{{\rm B\kern-.05em{\sc i\kern-.025em b}\kern-.08em
    T\kern-.1667em\lower.7ex\hbox{E}\kern-.125emX}}

\usepackage{amsthm}

\usepackage{tabularx}
\usepackage{makecell}
\usepackage{multirow}

\usepackage{amsmath}
\usepackage{amssymb}
\usepackage{bbm}
\usepackage{bm}
\usepackage{comment}
\usepackage[capitalize]{cleveref}

\usepackage{pgfplots}
\usepackage{float}
\usepackage{booktabs}

\pgfplotsset{compat=1.15}

\usepackage[bottom=1.1 in,top=0.75in,left=0.63in,right=0.63 in]{geometry}

\begin{document}

\title{Age and Power Minimization via Meta-Deep Reinforcement Learning in UAV Networks
}
\author{
	\IEEEauthorblockN{Sankani Sarathchandra, Eslam Eldeeb, Mohammad Shehab, Hirley Alves, Konstantin Mikhaylov, and Mohamed-Slim Alouini\\
	}
	\thanks{Sankani Sarathchandra, Eslam Eldeeb Hirley Alves and Konstantin Mikhaylov are with Centre for Wireless Communications (CWC), University of Oulu, Finland. 
                (e-mail: firstname.lastname@oulu.fi). 
                Mohammad Shehab and Mohamed-Slim Alouini are with CEMSE Division, King Abdullah University of Science and Technology (KAUST), Thuwal 23955-6900, Saudi Arabia (email: mohammad.shehab@kaust.edu.sa,slim.alouini@kaust.edu.sa).
        }
	\thanks{The work of S. Sarathchandra, E. Eldeeb, H. Alves and K. Mikhaylov was supported by 6G Flagship (Grant Number 369116) funded by the Research Council of Finland. The work of S. Sarathchandra and K. Mikhaylov was additionally supported by RCoF projects FIREMAN (348008) and MRAT-SAfeDrone (341111), respectively. 
        }
}
\maketitle

\begin{abstract}
Age-of-information (AoI) and transmission power are crucial performance metrics in low energy wireless networks, where information freshness is of paramount importance. This study examines a power-limited internet of things (IoT) network supported by a flying unmanned aerial vehicle (UAV) that collects data. Our aim is to optimize the UAV's flight trajectory and scheduling policy to minimize a varying AoI and transmission power combination. To tackle this variation, this paper proposes a meta-deep reinforcement learning (RL) approach that integrates deep Q-networks (DQNs) with model-agnostic meta-learning (MAML). DQNs determine optimal UAV decisions, while MAML enables scalability across varying objective functions. Numerical results indicate that the proposed algorithm converges faster and adapts to new objectives more effectively than traditional deep RL methods, achieving minimal AoI and transmission power overall.
\end{abstract}

\section{Introduction}\label{sec:introduction}
The future of wireless communication involves the deployment of large-scale and complex scenarios such as vehicle platooning, smart factories, and connected and autonomous vehicles (CAVs)~\cite{9762548,da2024distributed}. These applications often come with stringent quality of service (QoS) requirements, including ultra-low latency, high reliability, limited transmission power and information freshness~\cite{petar_perspective,mahmood2020white}. Age of information (AoI) is a metric  that measures the freshness of data, \textit{i.e.}, the time elapsed since the last received packet was generated.
Minimizing the AoI is a key challenge across a wide range of wireless communication applications~\cite{AoI_orig}, particularly for the uplink of low energy devices~\cite{10342862,9845353}.

Recently, unmanned aerial vehicles (UAVs) have been explored as flying base stations (BSs) to optimize AoI and transmission power~\cite{UAVs_tutorial}. UAVs offer remarkable flexibility by positioning themselves closer to the devices they serve, thus improving line-of-sight (LoS) probability and increasing the likelihood of successful transmission. Additionally, by reducing the distance between the UAV and the devices, energy consumption is minimized, leading to longer battery life for the devices. UAVs can optimize their flight trajectories and device scheduling policies to jointly minimize AoI and transmission power jointly~\cite{eldeeb2022multi}.

In this regard, future wireless networks, especially UAV-based networks, must incorporate intelligent decision-making algorithms. Machine learning and artificial intelligence (ML/AI) are pivotal for the success of $5$G and beyond and future $6$G networks, providing the necessary flexibility and scalability to manage large and complex wireless environments~\cite{10198239}. Among ML techniques, reinforcement learning (RL) stands out for its effectiveness in decision-making and control tasks. RL typically frames problems as Markov decision processes (MDPs), where an agent observes the current state of the environment, takes an action, transitions to a new state, and receives a reward that corresponds with the effectiveness of the chosen action in that state. Given the sequential nature of wireless networks, UAV trajectory planning can be framed as an MDP and solved using efficient RL algorithms to jointly minimize both the AoI and transmission power~\cite{chen2021deep,8714026}.

Although conventional RL has shown promising potential in UAV networks, it faces two main challenges. First, the UAV system is typically more complex and high-dimensional, particularly in terms of the size of the state and action spaces. Second, the policies designed for a specific network setup, defined by factors such as the number of devices, their locations, the channel model, and the traffic patterns of the applications they serve, are often not transferable to networks with different configurations or characteristics. To address these challenges, deep RL and meta-learning have emerged as two promising frameworks designed to overcome these limitations.

Deep RL has emerged as a promising solution to address the limitations of traditional RL. By combining deep neural networks with conventional RL techniques, deep RL offers an explicit approach to handling complex, high-dimensional environments. One notable deep RL algorithm is the deep Q-Network (DQN)~\cite{DQNs}, which revolutionized the RL field due to its ability to master complex video games without prior knowledge. DQNs use deep neural networks as function approximators to estimate the Q-function, which evaluates the quality of each action in a given state. Moreover, DQNs rely on off-policy learning, where experiences from different policies are used to improve the current policy. Deep RL, specifically DQN, has been proven to be highly effective in UAV networks for jointly optimizing AoI and transmission power~\cite{kim2024energy}.

In addition to deep RL, meta-learning is another class of algorithms that has gained significant traction in the wireless domain. Meta-learning, often referred to as \emph{learning to learn}, trains machine learning models to quickly adapt to new tasks with minimal data. One well-known meta-learning algorithm is model-agnostic meta-learning (MAML), which aims to identify a set of optimal initial model weights that can rapidly converge with just a few stochastic gradient descent (SGD) steps and a small amount of training data~\cite{FINN}. MAML leverages learning across different tasks to determine generalized initial weights that perform well across similar tasks. The key distinction between transfer learning and meta-learning is that transfer learning involves applying a pre-trained model from one task to another. In contrast, meta-learning focuses on training a model across multiple tasks to find optimal initial parameters, avoiding the need to train from scratch. Meta-learning offers significant scalability by enabling rapid adaptation to varying combinations of AoI and transmission power objectives without the extensive fine-tuning required by transfer learning, making it ideal for minimizing these metrics, especially when configuration changes, such as device locations or channel models.

\subsection{Literature Review}

Recent advancements in UAV-assisted wireless networks have focused on minimizing AoI to ensure real-time data freshness in dynamic applications. The use of UAVs for enhancing wireless communication coverage and data collection has been studied in various challenging environments in~\cite{7470933}. This adaptability enables UAVs to maintain real-time connectivity in areas without infrastructure and supporting applications. In~\cite{9151993}, the authors proposed a UAV trajectory optimization framework to minimize AoI in wireless sensor networks (WSNs) by balancing data transmission and flight times for timely data collection. Alternatively, a deep deterministic policy gradient (DDPG) algorithm was proposed in~\cite{9455139} to balance AoI, energy consumption, and data collection efficiency in UAV networks, highlighting the necessary trade-offs in real-time decision-making. This approach uses a multi-objective framework that allows the UAV to adjust its behavior based on priorities, optimizing performance across competing demands. 

Recent studies have explored RL and deep RL techniques to enhance UAV performance in wireless networks. A recent survey in~\cite{10283826} highlights the extensive applications of RL in multi-UAV wireless networks, discussing its potential for trajectory planning and dynamic resource allocation to support complex and collaborative missions. Authors in~\cite{9322234} developed a novel RL approach using a double deep Q-network (DDQN) to optimize UAV path planning for data collection in urban IoT networks, overcoming challenges such as limited flight time and the presence of obstacles. 
Building on existing reinforcement learning applications in wireless networks, authors in~\cite{8928091} went one step further. They proposed a deep RL-based trajectory planning algorithm to minimize the AoI in UAV-assisted IoT networks. By framing the trajectory planning as a MDP, their approach dynamically adjusts to unpredictable traffic patterns, demonstrating enhanced data freshness and robustness in real-time UAV operations. Another trajectory planning algorithm is discussed in~\cite{9507262} for energy consumption minimization in UAV-assisted WSNs using deep RL. This approach prioritizes energy-efficient UAV paths by dynamically adjusting the trajectory in response to network conditions, extending operational time, while maintaining effective data collection. The study in~\cite{9465671} focuses on using deep RL to optimize resource allocation, including UAV positioning and communication management, to enhance network performance and throughput in cooperative UAV systems. 

Meta-learning has shown significant promise in wireless communications by enabling fast adaptation to new environments with minimal data. The survey in~\cite{10477590} reviews various meta-learning techniques, emphasizing their ability to optimize wireless policies such as beamforming, resource allocation, and channel estimation, even when the underlying network conditions change dynamically. Authors in~\cite{loli2024meta} propose a meta-learning approach to optimize non-convex problems in large-scale wireless systems, highlighting how meta-learning can significantly reduce the complexity of conventional optimization algorithms. Meta-reinforcement learning (Meta-RL) leverages meta-learning techniques to boost the adaptability of RL models, enabling faster learning and improved performance in dynamic environments with limited data. Meta-RL can be utilized for trajectory optimization of UAVs as the meta-learning algorithm tunes the hyperparameters of an RL solution, allowing the UAV to adapt quickly to new environments~\cite{9322414}. Another meta-RL approach was introduced by authors in~\cite{10279524} to optimize unmanned aerial base stations trajectory planning. This method adapts quickly to new traffic patterns by transferring knowledge from previous configurations, reducing the number of training episodes needed to optimize its policy for new scenarios.

\subsection{Contributions}
This paper presents a meta-deep RL framework combining DQN with MAML for efficient adaptation to dynamic multi-objective optimization in UAV-aided IoT networks. The main contributions of this paper are summarized as follows:
\begin{itemize}
    \item We consider the problem of optimizing the UAV trajectory and its scheduling policy to jointly minimize the AoI and transmission power of low energy IoT devices. We solve the problem using DQN, a well-known deep RL algorithm, which estimates the optimum Q-function using a neural network.
    
    \item We propose combining DQN with MAML, a meta-learning algorithm, to ensure scalability when changing the network configurations.

    \item We consider a multi-objective problem that targets minimizing the weighted AoI and weighted power. In this problem, changing the weights of the objective elements is the source of variation in the objective function. Hence, we target fast adaptation for different objectives.

    \item Experimental simulations demonstrate that the proposed meta-deep RL algorithm outperforms the baseline RL model without MAML in terms of fast adaptation to new objectives, the need for few-shots of experiences, and the overall achievable AoI and transmission power.

\end{itemize}

\subsection{Outline}
The rest of the paper is organized as follows: Section~\ref{sec:sysmodel} introduces the system model. The proposed meta-RL algorithm is presented in Section~\ref{sec:Meta_RL}. Section~\ref{sec:results} evaluates the proposed algorithm through numerical analysis. Finally, conclusions and future work are discussed in Section~\ref{sec:conclusions}. Table~\ref{Notations} summarizes the list of notations used in the paper.




\section{System model and Problem Formulation}\label{sec:sysmodel}
Consider an $L \times L$ $2$D grid populated with stationary limited-power IoT devices (termed simply devices in what follows), represented by $\mathcal{D}=\{1,2,\cdots, D\}$.  Each device $d \in D$ is randomly deployed and positioned at a specific coordinate $c_d=(x_d,y_d)$ on the grid. We consider a UAV flying above this grid at a fixed altitude, whose task is to gather data from the devices. The UAV starts flying from an initial random position within the grid, and its movement is constrained, ensuring it does not egress from the grid's boundaries. The UAV can move in four cardinal directions (north, east, south, and west) or hover in its current position.

\begin{figure}[t!]
	\centering
	\includegraphics[width=0.95\columnwidth]{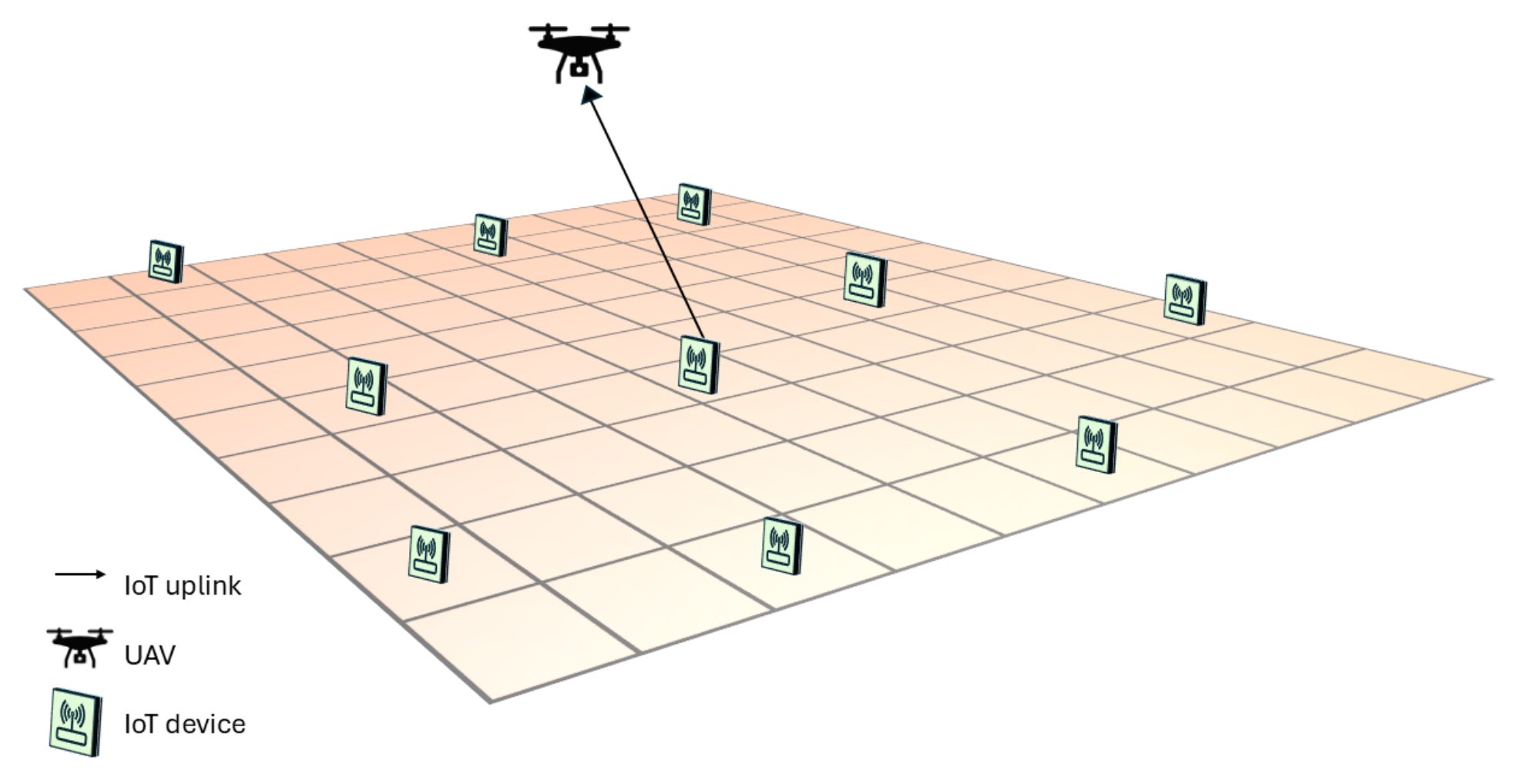}
	\caption{The considered system model operating within a grid environment, consists of a UAV that serves multiple, fixed located IoT devices with limited power.} 
	\label{Fig1}
\end{figure}

In this model, we consider an episodic environment, where an episode is segmented into discrete time intervals, $[t, 2\: t, \cdots]$, where $t$ specifies the duration of each time step the UAV moves from the center of a cell to the center of an adjacent cell, according to the selected movement direction, and receive information from a device. The position of the UAV at time $t$ is fully described by its $2$D projection $c_u = (x_u,y_u)$ 
on the plane and its altitude $h_u$\footnote{In this study, the UAV's orientation is assumed to have a negligible impact on the optimization objectives of AoI and transmission power.}. The system model is illustrated in Fig. \ref{Fig1}.  



The scheduling policy determines which device can transmit data during each time slot. Scheduling policy $w_u(t) = d$ indicates that device $d$ is set to transmit during time slot $t$. Only one device transmits per time slot to avoid interference. We assume a line-of-sight (LoS) communication between the devices and the UAV, where the channel gain between the UAV and device $d$ at time slot $t$ is given by~\cite{eldeeb2023age} 
\begin{align}
\label{LOS_eq}
    g_{u,d}(t) = \frac{g_0}{|h_u|^2+r_{u,d}^2(t)},
\end{align}
where $g_0$ is the channel gain at the reference distance of 1 m, $h_u$ is the altitude of the UAV, and $r_{u,d}(t)$ is the Euclidean distance between the device and the UAV at time $t$.  Hence, the transmit power $P_d$ of device $d$ at time $t$ is calculated as follows
\begin{align}
    P_d(t)\! &=\! \frac{(2^{\frac{M}{BW}}-1)\sigma^2}{g_{u,d}(t)}
    \!=\!\frac{(2^{\frac{M}{BW}}-1)\sigma^2}{g_0}\!\Bigg(r_{u,d}^2 (t) \!+\! h_u^2\!\Bigg),
\end{align}
where $M$ is the packet size of the received message, $BW$ is the signal bandwidth and $\sigma^2$ the noise power.

We use the AoI as a performance metric to quantify the freshness of data collected from IoT devices~\cite{8928091}. The AoI for a device $d$ at a given time $t$, denoted as $A_d(t)$, is defined as the time elapsed since the last successful time update received from device $d$ was generated. Assuming instant transmission, the evolution of AoI for device $d$ at the next time step $t+1$ is given by
\begin{equation}
\label{AOI_CALC}
	A_d(t+1) =
	\begin{cases}
		1, & \quad \text{if} \ 
        w_u(t) = d, \\
		\text{min}\{A_{max},A_{d}(t) + 1\}, & \quad \text{otherwise}, 
	\end{cases}
\end{equation}
where $A_{max}$ denotes the maximum practically allowed AoI in the model.
 

\subsection{Problem Formulation}

This problem aims to optimize the UAV trajectory and its scheduling policy to minimize the AoI and the power consumption of the IoT devices using a few shots of interactive data with the environment. The objective function varies depending on the trade-off between AoI and power consumption, which is controlled by a parameter to address different optimization priorities. 
We formulate the problem as
\begin{subequations}\label{P1}
	\begin{alignat}{2}
        \mathbf{P1:} \qquad &\underset{c_u(t),S(t)}{\min}       &\ \ \ & \frac{1}{T}\sum_{t=1}^T\sum_{d = 1}^{D}\ \omega A_d(t) + \frac{\lambda}{D} \sum_{d = 1}^{D} P_d(t),\label{P1:a}
	\ \\
	&\text{s.t.}   &      & 
        \quad e_i \leq E_{max}, \quad \forall i \in \{1, 2, \dots, I\}, \label{P1:b}\
	\end{alignat}
\end{subequations}
where $\lambda$ is a scaling factor that controls the trade-off between AoI and transmission power. $\omega$ denotes the  weight assigned to each device $d$ reflecting its relative importance of AoI. $I$ represents the number of tasks. A higher value of $\lambda$ increases the weight of minimizing the power consumption, while a lower value prioritizes the AoI. If $\lambda = 0$, the model focuses entirely on optimizing AoI without considering transmission power. Varying $\lambda$ generates different tasks during training, allowing meta-learning to train the model to adapt efficiently to these trade-offs.
The constraint \eqref{P1:b} ensures that the number of episodes $e_i$ used to train the task $i$ remains within the allowed few-shot budget $E_{max}$, promoting efficient learning with minimal data.

The optimization problem in~\eqref{P1:a} is a multi-objective non-linear optimization problem, as it targets minimizing two objectives: AoI, to ensure timely and updated information in the network, and transmit power, to conserve the limited energy resources of IoT devices.
The complexity of this optimization problem increases with the number of devices and variations in the trade-off parameter $\lambda$. This problem is likely NP-hard because it involves combination of discrete scheduling decisions and continuous trajectory optimization, along with the non-linear and multi-objective nature of the problem. These factors makes finding an exact solution computationally infeasible in large-scale scenarios. With the IoT network's dynamic nature and the UAV's need to adapt to varying environments and tasks, we propose a meta-based RL algorithm by combining meta-learning (see details in Section \ref{ssec:ML-maml}) with RL.

\begin{table}[t!]
\centering
\caption{Notations}
\label{Notations}
\begin{tabular}{c|c}
\hline

$A_d(t)$ & AoI of device $d$ at time step $t$ \\ 

$a(t)$ & Action of agent at time step $t$ \\ 

$B$ & Size of the replay buffer \\ 

$c_u(t)$ & Position of the UAV at time step $t$ \\ 

$d_o$ & Distance between centers of adjacent grid cells \\ 

$E$ & Number of episodes \\ 

$E_{max}$ & Few shot budget \\ 

$g_{u,d}(t)$ & Channel gain between UAV and device $d$  \\ 

$g_{0}$ & Channel gain at the reference distance of $1 \: m$  \\ 

$g_{t}$ & Return  \\ 

$I$ & Number of tasks \\ 

$I$ & Number of epochs \\ 

$\nabla_\theta\mathcal{L}_{\mathcal{T}_i}(\theta)$ & Gradient of the loss of task $i$  \\ 

$\mathcal{L}(\theta)$& Meta loss \\ 

$m_u$& Movement direction of the UAV \\ 

$P_d$ & Transmit power of device \\ 

$P(s(t+1) | s(t),a(t)$ & Transition probability \\ 

$Q(s(t),a(t))$ & Q-function at time step $t$  \\ 

$r_{u,d}$ & Euclidean distance between UAV and the device $d$ \\ 

$r(t)$ & Immediate reward at time step $t$\\ 
 
$s(t)$ & State of the environment at time step $t$ \\ 

$s(t+1)$ & State of the environment at next step of time step $t$ \\ 
 

$\omega$ & Importance weight for the AoI of the device $d$ \\ 

$w_u(t)$ & Scheduling policy of UAV at time step $t$   \\ 

$\alpha$ & Adaptation learning rate\\ 

$\beta$ & Meta-learning rate\\ 

$\gamma$ & Discount factor \\ 
 
$\theta$ & Initial weight of the task \\ 

$\theta'$ & Updated weight of the task \\ 

$\tau_i$ & Task $i$ \\ 

$\pi^*$ & Optimal policy of agent \\ 
 
$\lambda$ & Scaling factor for power consumption in the reward \\ 
 
$\epsilon$ & Probability of exploration \\ 
 
$1-\epsilon$ & Probability of exploitation \\ 

\hline
\end{tabular} 
\end{table}
\section{Background}\label{sec:Meta_RL}
In this section, we introduce deep reinforcement learning and meta-learning fundamentals. These materials will help introduce the proposed deep meta-RL algorithm presented in the next section.




\subsection{Markov Decision Processes Formulation}

 The RL problem is formulated as a Markov Decision Process (MDP), which is composed of the tuple $\langle s, a,r,p \rangle$, where $s$ is the state, $a$ is the action, $r$ denotes the reward function and $p$ is the state transition probability. At time instant $t$, the agent observes the current state $s(t)$ from the environment, selects an action $a(t)$ following a policy $\pi$, which describes the selected action at each particular state, receives an immediate reward $r(t)$ for selecting that action at that state, and transits to the next state $s(t+1)$ according to the transition probability $p(s(t+1)|s(t), a(t))$. The agent aims to find the optimum policy that maximizes the return $G(t)$, which is the accumulative discounted reward 
\begin{equation}
    G(t) = \sum_{k=0}^{\infty} \gamma^{k} r(t+k+1),
\end{equation}
where $\gamma$ is the discount factor that controls how long we care about future rewards compared to immediate rewards. Setting $\gamma \rightarrow 0$ means we focus on immediate rewards more than future rewards, whereas $\gamma \rightarrow 1$ considers future rewards in the return function. Next, we define each of these elements in the context of the formulated problem and the target is to optimize both UAV trajectory and its scheduling in order to minimize the AoI and transmission power.

\subsubsection{State space}
 The state space represents all the observations available to the agent at a given time $t$. The state space of the system at time slot $t$ is defined as $s(t) = (\boldsymbol{c_u}(t),\boldsymbol{A}(t))$ where $\boldsymbol{c_u}(t)$ is the current $2$D location of the UAV at time slot $t$, and $\boldsymbol{A}(t) = (A_1(t), A_2(t),..., A_D(t))$ is a vector that contains the individual AoI of each IoT device, while $A_d(t)\in \{1,2,..., A_{max}\}$. The cardinality of the state space is $|s|=2 + D$. 

\subsubsection{Action space}
The action space represents the set of available decisions that the agent can take based on the current state. The action space at time slot $t$ is defined as $a(t) = (w_u(t),m_u(t))$, where $w_u(t)$ is the scheduling policy of the UAV (\textit{i.e., $w_u(t) = d$} means that the UAV chooses device $d$ to serve at time $t$), and $m_u(t)$ is describes the movement direction of the UAV (\textit{i.e.,} north, south, east, west, or hovering). The number of possible actions is given by $D \times 5$.

\subsubsection{Reward function} 
The reward function is defined by the weighted 
average AoI values and energy consumption of all the devices for each action. A scaling factor, $\lambda$, is used to evaluate the trade-off between energy consumption relative to AoI. The overall goal of the UAV is to maximize its reward over time by making choices that efficiently balance these two objectives, keeping data up to date and minimizing energy use. Reward $r(t)$ at the time step $t$ is quantified by~\cite{eldeeb2023traffic}
\begin{equation}
\label{REW_CALC}
r(t) = -\frac{1}{D} \sum_{d = 1}^{D} \omega A_d(t) \: - \lambda \:  \sum_{d = 1}^{D} P_d.
\end{equation}
As the UAV explores different actions, it learns to optimize the reward by finding the most effective combination for minimizing AoI and conserving power.

\subsubsection{State transition probability} 
The transition between states relies on the two components of the state space. The AoI is updated according to \eqref{AOI_CALC}. The movement direction of the UAV at a given time slot is modeled as
\begin{equation} \label{eqn:directions}
	c_u(t+1)=
	\begin{cases}
		c_u(t)+(0,d_0), & \quad m_u(t)=\text{North}, \\
		c_u(t)-(0,d_0), & \quad m_u(t)=\text{South}, \\
		c_u(t)+(d_0,0), & \quad m_u(t)=\text{East}, \\
		c_u(t)-(d_0,0), & \quad m_u(t)=\text{West}, \\
		c_u(t), & \quad \text{Hovering}, \\
	\end{cases}
\end{equation}
where $d_0$ is the distance between the centers of two adjacent cells.

\subsection{Deep Reinforcement Learning}



The agent's objective is to learn the optimal policy $\pi^*$ that maximizes the accumulative discounted rewards. The Q-function $Q(s(t), a(t))$ maps the expected reward of being at a state $s(t)$, taking an action $a(t)$, and transiting to state $s(t+1)$. Hence, to find the optimal policy, we need to find the optimal Q-function $Q^*(s(t), a(t))$, which is achieved using a Q-learning algorithm that updates the current Q-function iteratively using the Bellman equation
\begin{align}
\label{BELLAMN_EQ}
 Q&\left(s\left(t\right),a\left(t\right)\right) = \:  Q\left(s\left(t\right),a\left(t\right)\right) + \nonumber\\
&\alpha \:  \left(r\left(t\right) +  \gamma \: \max_a Q\left(s\left(t+1\right),a\right)
-Q\left(s\left(t\right),a\left(t\right)\right)\right),
\end{align} 
where $\alpha$ is the learning rate and $\gamma \: Q\left(s\left(t+1\right),a\left(t+1\right)\right)$ is the discounted state-action value at time instant $t+1$. The agent employs an $\epsilon$-greedy exploration policy, where it primarily selects actions that maximize the Q-function with a probability $1-\epsilon$ and explores suboptimal actions with a small, yet dynamic probability $\epsilon$ to ensure sufficient exploration of the environment. Frequently, $\epsilon$ is set to a large value at the first episodes to ensure enough exploration and then decays with time \cite{ADB+17_DRLsurvey}.

To this end, Q-learning stands short in complex and large-dimension environments as it needs to visit many state-action pairs. Deep RL extends the traditional reinforcement learning methods with deep neural networks. DQN is one family of deep RL that estimates the Q-function using deep neural networks. In addition, DQNs utilize key techniques to stabilize the learning process by using a target network and replay memory buffer. Specifically, DQNs adopt two neural networks that work together to improve the learning process. The first network estimates the current Q-values based on the agent's actions and states. The second is the target network, which is only updated periodically, reducing the correlations between the current Q-values and the target values~\cite{DQNs}. The replay memory stores past experiences and samples them randomly, which breaks correlations between consecutive experiences, resulting in more efficient and stable learning, and allowing the agent to converge toward an optimal policy in complex, high-dimensional environments. Algorithm~\ref{alg1} describes the implementation of deep RL. 
\begin{figure*}[t!]
	\centering
	\includegraphics[width=1.0\textwidth]{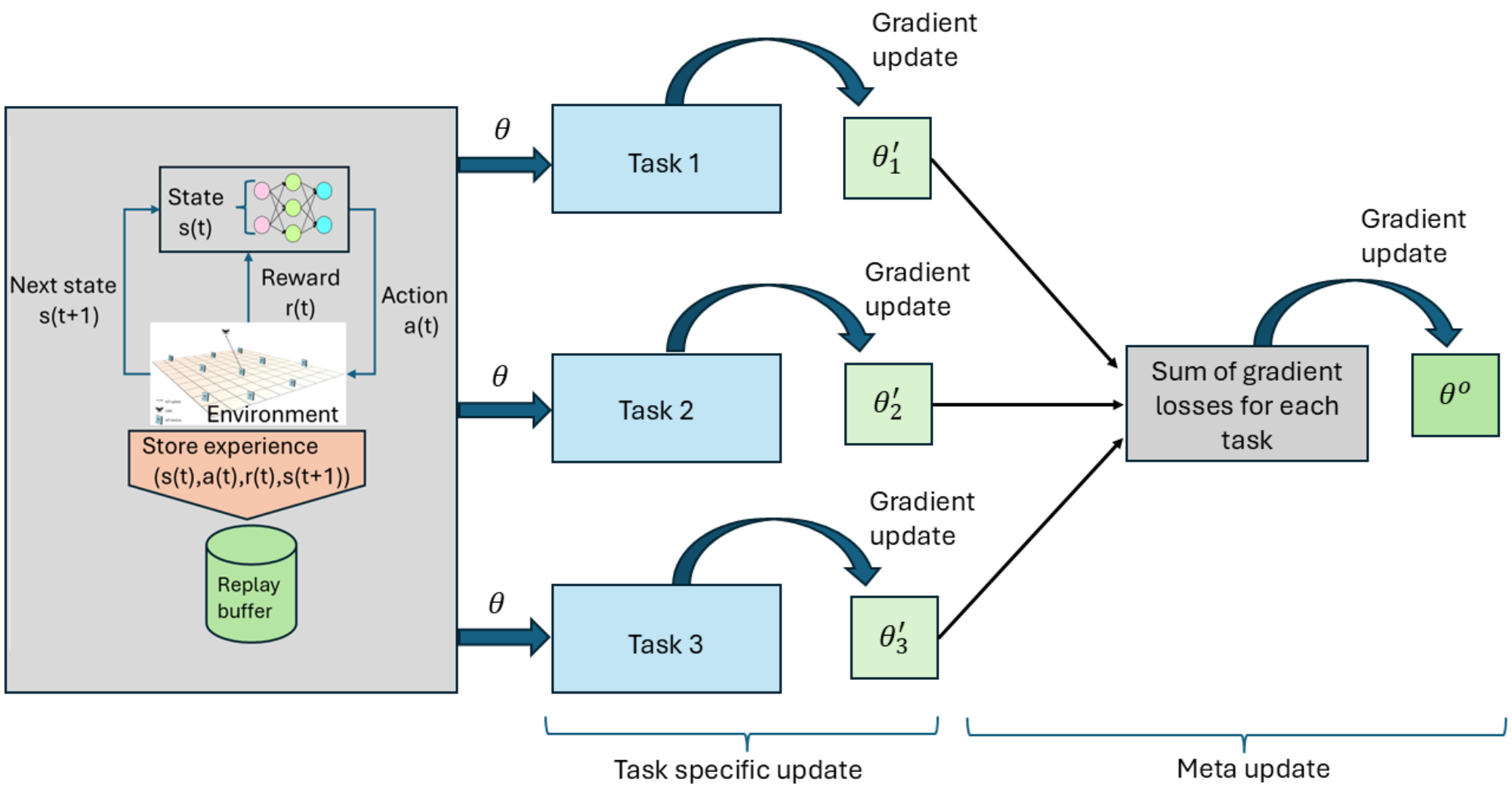}
        \captionsetup{justification=centerlast}
	\caption{A visualization of the proposed meta-RL algorithm using MAML and DQNs. The left block represents the interaction between the DQN agent and the environment, where state transitions are stored in the replay buffer for training. The shared initial weights ($\theta$) derived from the meta-learned initialization, are used for all tasks and fine-tuned through task-specific gradient updates, resulting in task-specific weights ($\theta^{'}_1$,$\theta^{'}_2$,$\theta^{'}_3$). The updated weights are used to calculate individual task losses, which are then aggregated into a meta-loss to compute the meta-update ($\theta^{0}$). We utilize learning across different tasks trained using DQN agents to reach initial weights that can adapt quickly to new unseen tasks.}
	\label{Fig2}
\end{figure*}

\begin{algorithm}[!t]
\SetAlgoLined

Initialize Q-network and target network weights  


Initialize replay buffer $B$

\For{\text{episode} $e$ in $\{1$,...,$E$\}}{
    
    \For{\text{time} $t$ in $\{1$,...,$T$\}}{
        Feed current state to the Q-network and select action $a(t)$ using $\epsilon$-greedy exploration

        Transit to next state $s(t+1)$ and observe immediate reward $r(t)$

        Store the experience $(s(t), a(t), r(t), s(t+1))$ in the replay buffer $B$
        
        Sample a mini-batch $b$ of transitions from $B$
            
            
        Update Q-network and target network
   }
}
\caption{Deep RL}
\label{alg1}  
\end{algorithm}

\subsection{Meta-Learning}\label{ssec:ML-maml}

Meta-learning is a framework designed to equip models with the capability to rapidly adapt to new tasks using minimal data. MAML builds on this concept by training models to learn the initial parameters that can be optimized for fast adaptation, allowing standard models to quickly adjust to new tasks with minimal updates~\cite{eldeeb2024semanticmetasplitlearningtinyml}. Utilizing learning across functions, we can find the initial parameters (model weights) $\theta$ that can reach convergence through a few SGD steps on new unseen tasks. To implement the adaptation and evaluation process during training, each task is divided into a support and query set. The support set is used to fine-tune the model to a specific task, while the query set evaluates the performance of the adapted model on the same task.

First, the weights are randomly initialized and assigned to each task. Then, we update the weights for each task as follows
\begin{equation}
\label{Task_update}
\theta^{'}_i = \theta - \alpha\nabla_\theta \mathcal{L}_{\mathcal{T}_i}(\theta),
\end{equation}
where $\alpha$ is the adaptation learning rate, $\mathcal{L}_{\mathcal{T}_i}(\theta)$ is a chosen task-specific loss function and $\nabla_\theta\mathcal{L}_{\mathcal{T}_i}(\theta)$ is the gradient of a chosen loss function for task $\mathcal{T}_i$ 
After adapting to individual tasks, we calculate the individual losses using the adapted weights for each task
\begin{equation}
\label{meta_loss}
\mathcal{L}(\theta)= \sum_{\mathcal{T}_i}\mathcal{L}_{\mathcal{T}_i}(\theta^{'}_i),
\end{equation}
where $\mathcal{L}(\theta)$ is called meta-loss, which is determined by the summation of all losses overall tasks. Then, the initial weights are updated using the meta-loss as follows~\cite{FINN}
\begin{equation}
\label{meta_update}
\theta \leftarrow \theta - \beta\nabla_{\theta} 
\mathcal{L}(\theta).
\end{equation}
Here, $\beta$  is the meta-learning rate. 
This process ensures that the model learns an initialization that allows it to adapt to new tasks rapidly. The MAML algorithm is summarized in Algorithm~\ref{alg2} and the visual representation of MAML is depicted in Fig. \ref{Fig2}

\begin{algorithm}[!t]
\SetAlgoLined

Initialize neural network weights $\theta$

\For{\text{epoch} $n$ in $\{1$,...,$N$\}}{

    Sample a batch of tasks $\mathcal{T}_i$ 

    \For{\text{task} $\mathcal{T}_i$ in batch}{

        
        Compute task-specific loss $L_{\mathcal{T}_i}(\theta)$ 
        and perform gradient descent to update the initial weights $\theta_i'$ for each task using~\eqref{Task_update}
       
        
        Compute loss for each task $L_{\mathcal{T}_i}(\theta_i')$ using the updated task-specific weights $\theta_i'$
    }

    Compute meta losses in~\eqref{meta_loss}

    Update the initial weights $\theta$ using~\eqref{meta_update} 
}
\caption{Meta learning (MAML)}
\label{alg2}  
\end{algorithm}

In the meta-testing phase, the model is applied to new, unforeseen tasks with the parameters it learned during meta-training. For each new task, a small batch of data (i.e, few shots) is used to fine-tune these initial parameters by performing a few gradient descent steps. The key goal is to assess how well the model can quickly adjust its parameters to the new task rather than training the model from scratch. The performance is measured by evaluating the fine-tuned model on a separate query set for the new task. This process ensures that the learned initialization is versatile and can be generalized across functions, highlighting the model's ability to quickly adapt and perform well in diverse environments.






\section{The proposed Meta-RL}
In this work, we propose a meta-RL approach where the UAV interacts with diverse tasks, which are generated by varying the trade-off parameter, $\lambda$ in the reward function, which controls the trade-off between AoI and transmission power and network configurations. 
Each corresponds to a different environment with a different objective. We target utilizing learning across tasks to adapt quickly to new unseen tasks. We adopt DQNs to optimize the UAV policy for each task and MAML to find the optimal Q-network initial weights that enable fast adaptation. Task-specific update~\eqref{Task_update} involves optimizing the model parameters for a specific environment by minimizing the task-specific loss using gradient descent. Task specific loss is modeled as
\begin{multline}
\label{task_spec_loss}
\mathcal{L}_{\mathcal{T}_i}\left(\theta\right) = 
\mathbb{E}_{b} \Bigg[ \Big( r\left(t\right) +
\gamma \max_{a\left(t+1\right)} Q\left(s\left(t+1\right), a\left(t+1\right); \theta\right) \\
- Q\left(s\left(t\right), a\left(t\right); \theta\right) \Big)^2 \Bigg],
\end{multline}
where $\mathbb{E}_{b}[.]$ denotes the  expectation taken over a mini-batch $b$ of transitions sampled from the replay buffer, with each transition consisting of state, action, reward, and next-state tuples. After adaptation, the updated parameters $\theta_i^{'}$ are evaluated by calculating the loss $\mathcal{L}_{\mathcal{T}_i}(\theta_i^{'})$
\begin{multline}
\label{loss_each_task}
\mathcal{L}_{\mathcal{T}_i}\left(\theta_i^{'}\right) = 
\mathbb{E}_{b} \Bigg[ \Big( r\left(t\right) +
\gamma \max_{a'} Q\left(s\left(t+1\right), a\left(t+1\right); \theta_i^{'}\right) 
 \\
- Q\left(s\left(t\right), a\left(t\right); \theta_i^{'}\right) \Big)^2 \Bigg].
\end{multline}
Then, the global-level update optimizes the shared parameters, which are the weights of the Q-network, across all tasks~\eqref{meta_update}, ensuring that these parameters are globaly optimized to a high extent to enable quick adaptation to new tasks.


During meta-testing, the UAV is introduced to unseen tasks where the environmental conditions differ from those encountered during training. The UAV utilizes the updated initial weights of the Q-network, learned during meta-training, as a starting point and adapts to these new tasks by performing a few gradient updates through the task-specific loss function. The model is tested in different environments where the positions of devices and $\lambda$ values vary, ensuring that the UAV's ability to adapt to diverse conditions is thoroughly evaluated. The proposed meta-RL approach is summarized in Algorithm~\ref{alg3}.

\begin{algorithm}[!t]
\SetAlgoLined



Initialize Q-network weights $\theta$

Initialize a replay buffer for each meta-training and meta-testing task

\For{\text{epoch} $n$ in $\{1$,...,$N$\}}{

    \For{each meta-training tasks $\tau_i$, $i$ in $\{1$,...,$I$\}}{        
        
        \For{\text{episode} $e$ in $\{1$,...,$E$\}}{
        
            
            \For{\text{time} $t$ in $\{1$,...,$T$\}}{

                Select an action using the current policy and $\epsilon$-greedy strategy

                Store the transition $(s(t), a(t), r(t), s(t+1))$ in the corresponding replay buffer

                Sample a mini-batch $b$ from the corresponding replay buffer $B$

                Compute task specific loss using~\eqref{task_spec_loss}
                
                Perform task-specific gradient updates to update the Q-network parameters $\theta_i'$ using~\eqref{Task_update}
                
            }
        }
        
        \For{each episode}{
        
            
            \For{\text{time} $t$ in $\{1$,...,$T$\}}{
            
            Calculate the loss using the updated Q-network $\mathcal{L}_{\mathcal{T}_i}(\theta^{'}_i)$ using~\eqref{loss_each_task}
            }
        }
    }

    Compute meta losses in~\eqref{meta_loss}
    
    Update the initial weights $\theta$ using~\eqref{meta_update}

}
Test the optimized policy, $\pi^*$  on new meta-testing tasks after convergence 
\caption{Meta-RL}
\label{alg3}
\end{algorithm}






\section{Experimental Analysis}\label{sec:results}

In this section, we discuss the simulation results to demonstrate the effectiveness of the proposed Meta-RL algorithm and compare it with the baseline RL model with random Q-network initialization. The environment is described by a grid world of $1000$ m $\times$ $1000$ m, which is divided into $10\times 10$ cells. We consider a UAV serving $5$ devices and $10$ devices, respectively. The Q-network consists of three layers with two hidden layers of $256$ neurons. We adopt ReLU as the activation function and Adam optimizer to update the neural network weights. The replay buffer has a size of size $100000$. Meta-training comprises $500$ epochs, which are trained using the Pytorch framework on the NVIDIA Tesla V100 GPU. The simulation parameters are defined in Table \ref{tab:uav}.

\begin{table}[t!]
\centering
\caption{simulation parameters}
\label{tab:uav}
\begin{tabular}{c|c}
\toprule
\textbf{Parameter} & \textbf{Value} \\  \midrule

Channel gain at $1$ m $g_0$ & $30$ dB\\
Bandwidth $BW$ & $1$ MHz \\
Height of the UAV trajectory $h_u$ & $100$ m \\
Packet size $M$ & $5$ Mb \\
Noise power $\sigma^2$ & $-100$ dBm \\
Maximum allowed $A_{\text{max}}$ & $30$ \\
Adaptation learning rate $\alpha$ & $0.0001$ \\
Meta-Learning Rate $\beta$ & $0.0001$\\
Discount Factor $\gamma$ &  $0.99$\\

\bottomrule
\end{tabular}
\end{table}

\begin{figure*}[t]
    \centering
    \subfloat[Meta losses\label{Meta_Losses}]{\includegraphics[width=0.33\textwidth]{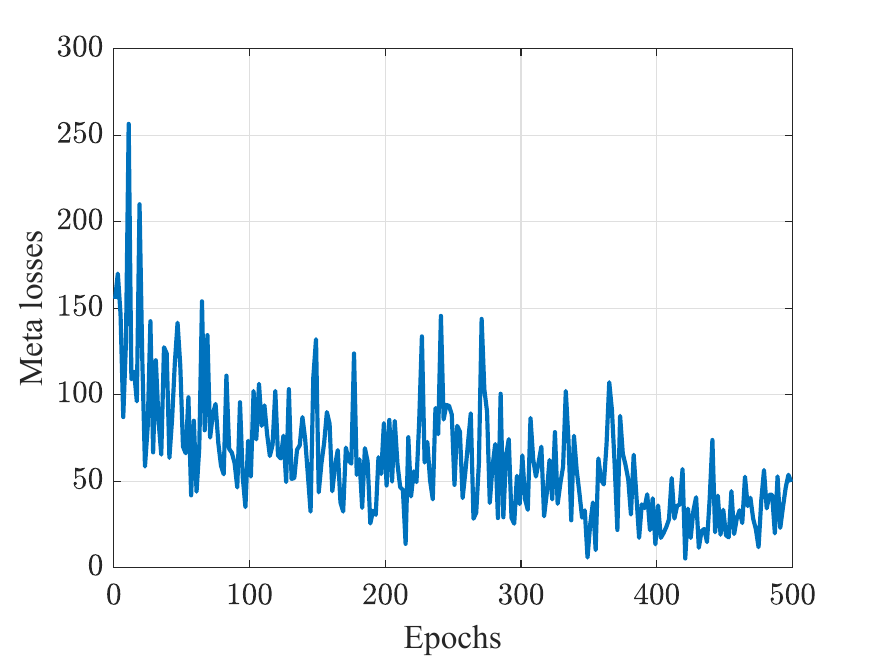}}
    \hskip -1.9ex
    \subfloat[$5$ devices\label{Meta_Train_5}]{\includegraphics[width=0.33\textwidth]{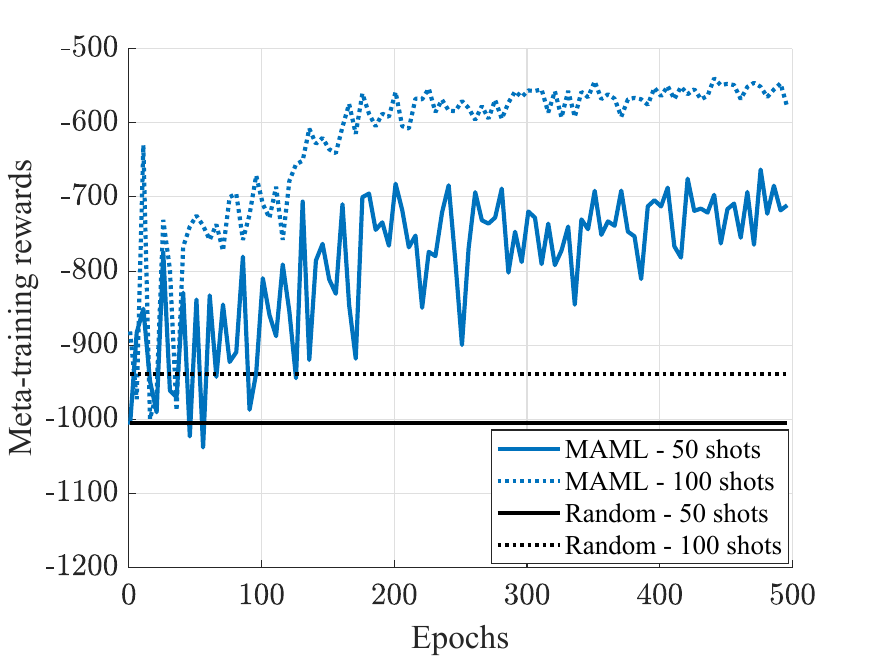}}
    \hskip -1.9ex
    \subfloat[$10$ devices\label{Meta_Train_10}]{\includegraphics[width=0.33\textwidth]{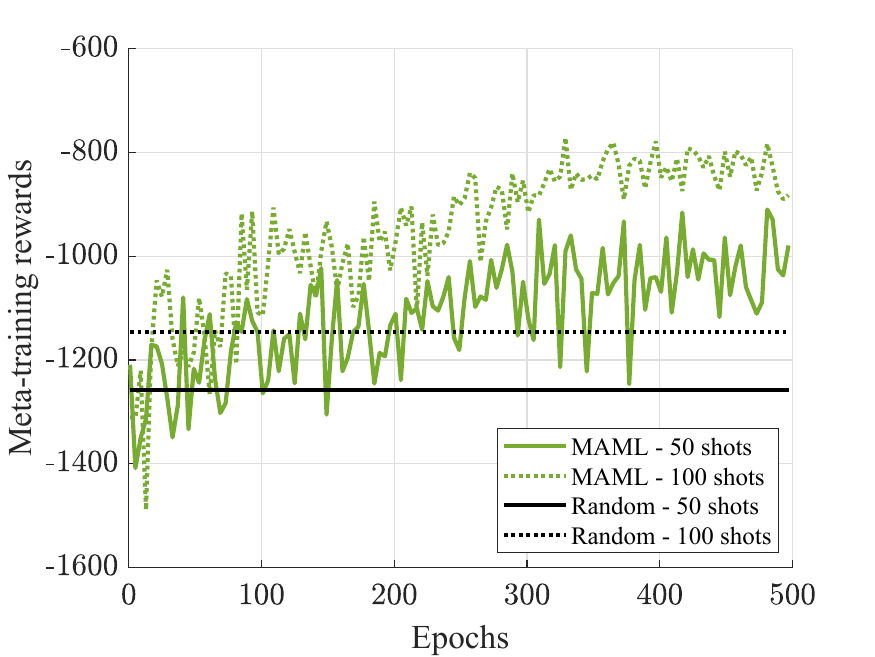}}
    \hskip -1.9ex
    \caption{Meta-training performance: (a) the convergence of the meta-losses, (b) the convergence of the rewards as a function of meta-training epochs tested over $11$ different test environments, each with $5$ devices, and (c) the convergence of the rewards as a function of meta-training epochs tested over $11$ different test environments, each with $10$ devices.}
    \label{Rate} 
\end{figure*}

\begin{figure}[t!]
    \centering    \includegraphics[width=1\columnwidth,trim={0cm 0 0cm 0},clip]{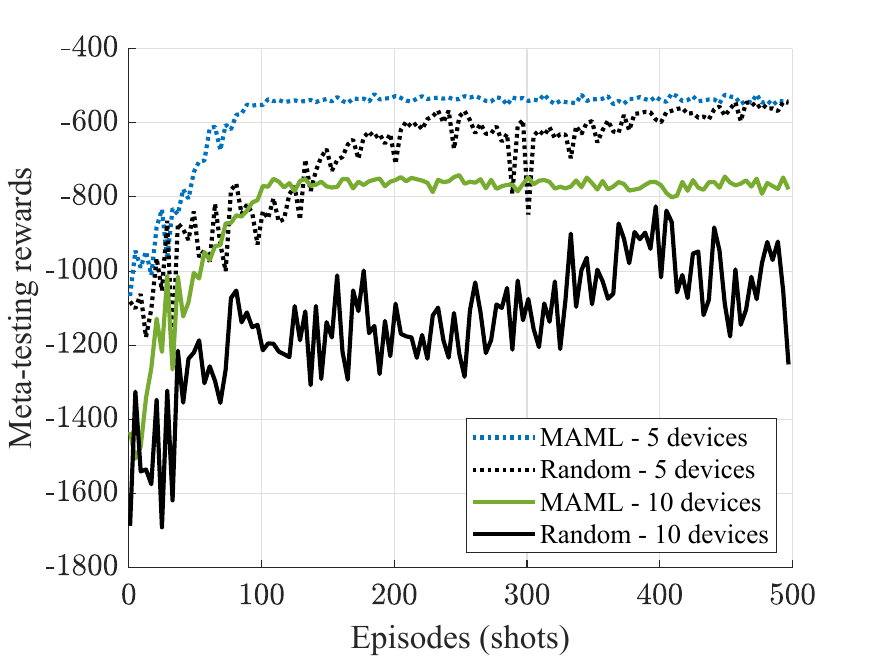} 
    \caption{Meta-testing convergence compared to random weights initialization with an environment with $\lambda = 300$.}
    \label{Meta_testing}
\end{figure}

We define $10$ different meta-training objective functions with varying $\lambda$ values, 
which adjusts the priority assigned for power consumption in the reward function allowing the model to learn and adapt to various setups. We consider equal importance of AoI across all devices, \emph{i.e.} $\omega = 1$.  Fig.~\ref{Rate} presents the model performance evaluation during meta-training. Fig.~\ref{Meta_Losses} highlights meta-training performance in terms of meta losses as a function of several training epochs. A batch of $6$ environments at a time is randomly sampled from the total set of training environments, where the meta-losses are averaged over each batch. We can notice that the convergence of meta-losses indicates that the model's ability to generalize across the training tasks is improving. Moreover, Fig.~\ref{Meta_Train_5} depicts the training convergence for environments with $5$ devices as a function of meta-training epochs, where each epoch comprises $100$ and $50$ episodes (shots), respectively. The model performance improves consistently as the number of shots increases, enabling the model to find the optimum policies. In addition, using MAML outperforms random initialization, which records bad convergence in both $50$ and $100$ shots cases. Similarly, Fig.~\ref{Meta_Train_10} demonstrates the effect of the number of shots on convergence in environments with $10$ devices, where a larger number of iterations consistently achieve higher rewards. Again and as concluded from Fig.~\ref{Meta_Train_5}, MAML outperforms random initialization regardless of the used number of shots.

\begin{figure}[t!]
    \centering    \includegraphics[width=1\columnwidth,trim={0cm 0 0cm 0},clip]{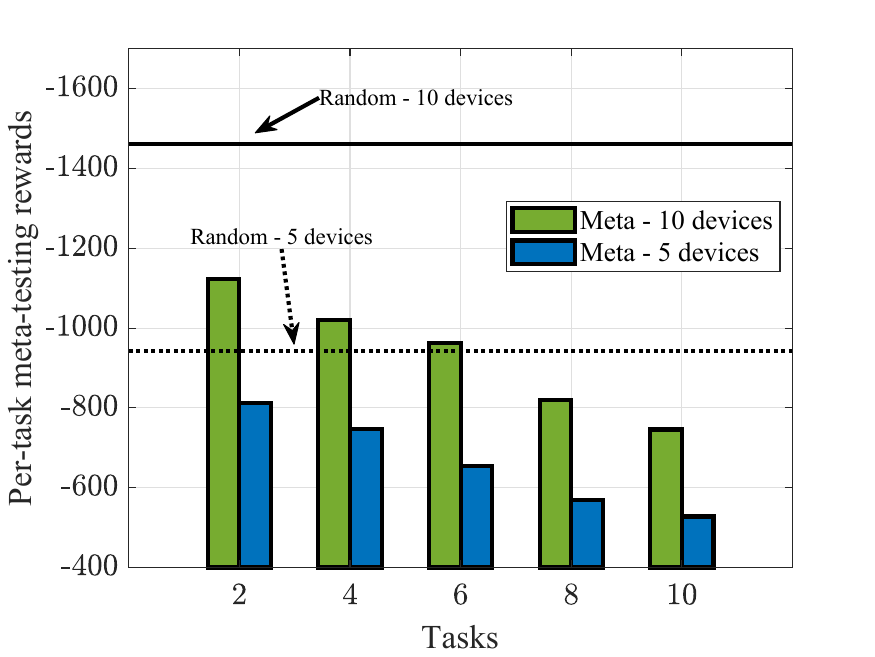} 
    \caption{Achieved rewards after $30$ episodes tested over $11$ different test environments.}
    \label{Meta_Tasks}
\end{figure}

During meta-testing, $11$ different test environments with different $\lambda$ values are used to assess how well the model can generalize. Fig.~\ref{Meta_testing} depicts the meta-testing rewards over $500$ episodes for both MAML-based  Q-network initialization and random Q-network initialization, setting $\lambda =300$ in an environment with $5$ devices and $10$ devices, respectively. The MAML-based initialization significantly outperforms the random initialization in both configurations, achieving convergence after around $100$ episodes, demonstrating that the model quickly adapts to the test environment and stabilizes its performance, using minimal fine-tuning. This rapid convergence during meta-testing reflects the effectiveness of the MAML initialization, allowing the model to generalize well to new environments after only a few adaptation steps. In contrast, the random initialization requires the whole $500$ episodes to reach convergence (sub-convergence) in both scenarios.

\begin{figure}[t!]
    \centering    \includegraphics[width=1\columnwidth,trim={0cm 0 0cm 0},clip]{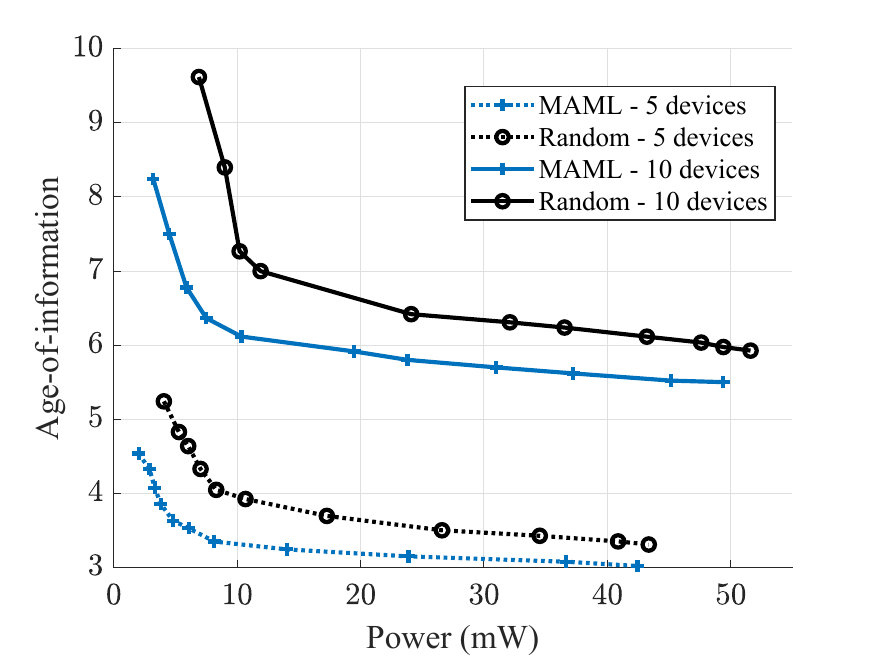} 
    \caption{Testing the proposed meta-RL algorithm over $11$ different environments (with different $\lambda$ values) regarding the achieved AoI and transmission power.}
    \label{Ach_Reg}
\end{figure}

Fig.~\ref{Meta_Tasks} displays the meta-testing rewards using $30$ episodes (shots), evaluated across $11$ test environments, as a function of the number of meta-training tasks. As the number of sampled training tasks increases, the rewards generally improve. In particular, in the $5$-devices scenario, the model consistently achieves better rewards even with training only $2$ tasks. Similarly, the $10$-devices scenario needs at least $6$ training tasks to find the initial weights that enable good convergence using only $30$ episodes. In contrast, the performance of the random initialization is significantly worse than the Meta-RL model in both scenarios. This highlights the importance of experiencing many tasks for better scalability as the Meta-RL model gains exposure to various environments, allowing it to learn more diverse patterns. As a result, the model develops a stronger generalization ability, enabling it to adapt more effectively to unseen test environments. The model fine-tunes its initialization by training on a broader set of tasks, leading to higher performance during meta-testing.

Fig.~\ref{Ach_Reg} demonstrates the achieved AoI and transmission power of the proposed meta-RL algorithm compared to random Q-network initialization while changing the value of $\lambda$. High age and low power values correspond to a small $\lambda$ and vice versa. The proposed algorithm outperforms the random initialization, achieving lower combined AoI and transmission power in the $5$-devices and $10$-devices scenarios for all $\lambda$ values. Note that the $5$-devices scenario is set in a lower region than the $10$-devices scenario due to the effect of the number of devices on the upper bounded AoI and transmission power. In addition, random initialization can reach the same regions with training more episodes. Therefore, the proposed meta-RL requires less converging time, saving time and computational power.



\section{Conclusions}\label{sec:conclusions} 

In this paper, we introduced a meta-deep RL algorithm to optimize the UAV's trajectory and scheduling policy to minimize AoI and transmission power. We first developed a DQN algorithm to determine the optimal UAV policy. Then, we integrated this DQN with MAML to identify the best initial weights of the Q-network, enabling rapid adaptation to dynamic networks with varying objectives. Simulation results highlighted the advantages of using MAML over conventional deep RL methods, demonstrating faster convergence within a few training episodes. Additionally, the proposed approach achieved the lowest combined AoI and transmission power for different objective functions. Deploying the proposed algorithm in a massive network served via multiple UAVs using multi-agent meta-RL (MAMRL) is left for the future.

\appendices 

\bibliographystyle{IEEEtran}
\bibliography{references}
\end{document}